\pdfoutput=1

\documentclass[11pt,a4paper]{article}
\usepackage[hyperref]{emnlp2020}
\usepackage{times}
\usepackage{latexsym}
\usepackage{amssymb}
\usepackage{amsmath}
\usepackage{xspace} %% si tu mets pas ça, ca marche pas le coup des espaces man :)
\usepackage{arydshln}

\usepackage{url}
\usepackage{xspace} %% si tu mets pas ça, ca marche pas le coup des espaces man :)
\usepackage{tablefootnote}

\usepackage{booktabs}

\usepackage{lingmacros}

    \usepackage{enumitem}
    \setlist{nolistsep}
\usepackage{microtype}

\usepackage{graphicx}
\newcommand{\mbert}{\mbox{mBERT}\xspace}
\newcommand{\random}{\mbox{Rand.}\xspace}

\newcommand{\mbertpos}{\mbox{mBERT+\textsc{POS}}\xspace}
\newcommand{\mberttask}{\mbox{mBERT+\textsc{Task}}\xspace}
\newcommand{\mbertparse}{\mbox{mBERT+\textsc{Parse}}\xspace}

\newcommand{\mbertadaptparse}{\mbox{mBERT+MLM+\textsc{Parse}}\xspace}
\newcommand{\mbertadapttask}{\mbox{mBERT+MLM+\textsc{Task}}\xspace}

\newcommand{\mbertadapt}{\mbox{mBERT+MLM}\xspace}

\newcommand{\camembert}{\mbox{CamemBERT}\xspace}

\newcommand{\arz}{\textit{Narabizi}\xspace}

\newcommand{\dummy}{{Majority Class}\xspace}
\newcommand{\stanford}{{StanfordNLP}\xspace}

\usepackage{enumitem}
\setlist{nolistsep}
\usepackage{microtype}

\usepackage{titlesec}
\titlespacing{\paragraph}{%
  0pt}{%              left margin
  0.1\baselineskip}{% space before (vertical)
  1em}%      

\titlespacing{\subsection}{0pt}{\parskip}{-\parskip}
\titlespacing{\subsubsection}{0pt}{\parskip}{-\parskip}

\usepackage{soulutf8}
\aclfinalcopy
\iffalse % set to iftrue for camera ready version, iffalse to display draft comments
    \usepackage[final]{proofing}
  \renewcommand\hl[1]{{#1}}  %% to remove the highlith
   {\draftnote{\red{#2}}}
   \newcommand\redHL[1]{}
  \newcommand\todo[1]{}
\else  % the stuff below is for the draft mode (hl, comments and so)
\usepackage[draft]{proofing}

\newcommand\red[1]{{\textbf{\textcolor{red}{#1}}}}

\let\oldred\red
\renewcommand\red[1]{{\bf \oldred{{#1}}}}

 \newcommand\redHL[1]{\red{\hl{#1}}}
\let\olddraftnote\draftnote
\renewcommand\draftnote[1]{\olddraftnote{\red{#1}}}

\fi

\title{Can Multilingual Language Models Transfer to an Unseen
  Dialect?\\ 
A Case Study on North African \textit{Arabizi}}

\author{Benjamin Muller \quad Beno\^{\i}t Sagot \quad Djam\'e Seddah \\
  Inria, Paris, France \\
  \texttt{firstname.lastname@inria.fr} \\}

\date{}

\begin{document}
\maketitle
\begin{abstract}
\iffalse
  Building natural language processing systems for non standardized
  and low resource languages is a difficult challenge. The recent
  success of large-scale multilingual pretrained language models
  provides new modeling tools to tackle this.  In this work, we study
  the ability of multilingual language models to process an unseen
  dialect with a high degree of variability. We take user-generated Algerian Arabic as our case
  study, a dialectal variety of Arabic with frequent
  code-mixing with French and written in Arabizi, a non-standardized
  transliteration of Arabic to Latin script.  We focus on two
  tasks, part-of-speech tagging and dependency parsing.  We show in
  multiple scenarios that multilingual language models are able to
  transfer to such an unseen dialect, specifically in two extreme
  cases: (i)~across scripts, using Modern Standard Arabic as a source
  language, and (ii)~from a distantly related language, unseen during
  pretraining, namely Maltese. 

 \else

 Building natural language processing systems for non standardized and
 low resource languages is a difficult challenge.  The recent success
 of large-scale multilingual pretrained language models provides new
 modeling tools to tackle this. In this work, we study the ability of
 multilingual language models to process an unseen dialect. We take user generated North-African Arabic as our case study, %We take user-generated content of North African Arabic as our case study, 
 a resource-poor dialectal variety of Arabic with frequent code-mixing with French and written in {\em Arabizi}, a
 non-standardized transliteration of Arabic to Latin script.
 Focusing on two tasks, part-of-speech tagging and dependency parsing,
we show in zero-shot and unsupervised adaptation scenarios 
that multilingual language models are able to transfer to such an
unseen dialect, specifically in two extreme cases: (i)~across scripts,
using Modern Standard Arabic as a source language, and (ii)~from a
distantly related language, unseen during pretraining, namely Maltese.
Our results constitute the first successful transfer
experiments on this dialect, paving thus the way for the development of an NLP ecosystem for resource-scarce, non-standardized and highly variable vernacular languages.
\iffalse
%% rire :) j'ai du indenter ces iffalse pour comprendre :)
for \iffalse {\em creole}-like,\fi{}resource-scarce and non standardized language \iffalse\else varieties.%, even when they exhibit a high degree of code-switching
\fi
\fi

\fi

% ** ranging from zero-shot cross-lingual transfer to low-resource scenarios and unsupervised domain adaptation,**

\iffalse
Our results show that even th
We also study the impact of using
  French as a source language to account for the code-switching
  component of our target language, and compare it with other,
  purposely less relevant
  languages. 

\fi
%Ranging from zero-shot cross-lingual transfer to low resource scenarios, our experiments shed light on the transfer ability and limit of pretrained multilingual models on unseen languages. %...
%Our contribution is twofold. We show that mBERT on Part-of-Speech tagging and dependency parsing outperforms non-contextual models despite not belonging to the pretraining corpora. We analyze extensively the properties of Arabizi that are successfully used by BERT and the one that are missed. We finally show that mBERT can be adapted in an unsupervised way to Arabizi. 
%Algerian Arabizi shows a high level of code-mixing with French as well as a high degree of variability across speakers. 
% POS Parsing 

%\textit{(unsup impact OR fact from detailed analysis OR unup better adaptation)}
% Should we do more experiments with other low ressource languages : North Sami, ... 
% SHOULD do on Arabic also ! maybe it will transfer ! 
% Can BERT transfer from transliteration
 
\end{abstract}

%% \section{Introduction}
%% under review par djamé

\section{Introduction}

Accurately modeling low resource and non-standardized languages exhibiting a high degree of variation is extremely challenging. Recent releases of multilingual language models trained on large corpora \cite{devlin2018bert,lample2019cross} provide an interesting opportunity to address this challenge in new ways. We frame our work as a \textit{cross-lingual transfer learning} analysis; we study the capacity of a system trained as a language model on a source set of languages to transfer to a target language and task.
%Indeed, by modelling in a single model such a large number of languages, we hope to capture cross lingual regularities. 
%We refer as \textit{unseen} any language or dialect that does not belong to the pre-training corpora 
  %that can be efficiently transferred to unseen languages. 
More precisely, we investigate the ability of multilingual language models to process a language that is absent from their pre-training set. For brevity, we  simply refer to such languages as \textit{unseen}. %Our study addresses two distinct scenarios: \textbf{zero-shot transfer} (no target data) and \textbf{unsupervised adaptation} (raw target data only). 
%\draftremove{By analyzing in detail the cross-lingual performance, we show that multilingual models are able to transfer to unseen, highly variable data.} %% déplacé plus loin

%our experiments shed light on the transfer abilities of multilingual models to unseen languages. 

%the question we ask is the following:
%\begin{itemize}
%    \item How does a multilingual pretrained language model can perform on an unseen dialect ?
%\end{itemize}

%\begin{enumerate}
%    \item How well does a multilingual pretrained network \underline{model}  an unseen language? (no tuning)
%    \item How well can a multilingual pretrained network \underline{learn}  an unseen language? (fine-tuning)
%\end{enumerate}
Our work focuses on the multilingual version of BERT (\mbert) \cite{devlin2018bert}. % trained on 104 languages of Wikipedia data. %We choose it because, up to this day,  \mbert is the available model that has been trained on the largest number of languages. 
The cross-lingual modeling ability of \mbert has been recently studied by \newcite{pires2019multilingual}, who show that cross-lingual transfer is very efficient between pretrained languages. In our work, we address a different and more challenging question: can \mbert transfer to an unseen and non-standardized dialect?\\%we analyse the ability of \mbert to model and learn an unseen dialect. \\
We take North-African Arabizi, hereafter \textit{\arz}, as our case study. We define \arz as 
%\cite{farrag2012arabizi}\draftnote{C'est pas la bonne ref. Elle parle de l'arabizi en général, pas du \arz - ds ca a été fixé ça ? : J'ai pas trouvé de meilleur ref. je voulais juste un ref. qui explique ce qu'est la romanization}
the Arabic dialect spoken in Algeria, found ubiquitously on social media and written in Latin script, although with no standard spelling and no standard transliteration of Arabic letters. It is a non-standardized dialect and shows a high degree of code-mixing with French \cite{amazouz2019addressing}. This makes \arz highly variable across users and therefore very challenging for Natural Language Processing.%{In addition to geographical variations, these points make of \arz a challenge for Natural Language Processing}.

% (NLP). %very appropriate to push the transfer abilities of pre-trained multilingual models. 

%It is non-standard, as there are no standard writing rules \bm{clear enough?}   and it shows a high degree of code-mixing \cite{Muysken} with French. 
%These two properties make it extremely challenging for Natural Language Processing and very appropriate to push the modeling abilities of pre-trained multilingual models. 
For our experiments, we use the  \arz raw corpus and treebank  recently released by \citet{seddah2020} and focus on two tasks, namely part-of-speech (POS) tagging and dependency parsing. %do fine-tuning and to evaluate our models on \arz Part-of-Speech (POS) tagging.  %which includes Part-of-Speech (POS) tags for 1200 sentences and 40  thousands raw sentences. 
After a detailed cross-lingual performance analysis, our results show that multilingual models are able to transfer to unseen, highly variable data.
More precisely, we make the following contributions:
\begin{itemize}
    \item We push the zero-shot cross-lingual abilities of \mbert to the extreme and show that it can transfer to unseen \arz in POS tagging and parsing, even when the source is another unseen and related language such as Maltese %\arz in Part-Of-Speech tagging, 
    %also absent from the pre-training corpora. 
    %\item We show that unsupervised fine-tuning, on raw \arz data impacts positively the zero-shot cross lingual and low resource scenarios, demonstrating the adaptability of multilingual language modflaels on non-standard and code-mixed data.
    \item By running comparison across source languages and diverse BERT models, we demonstrate that \mbert is using %\hl{making use}\draftnote{maladroit -ds} of 
    its multilingual representations to process \arz.
    \item We show the positive impact of unsupervised fine-tuning on cross-lingual transfer and demonstrate its ability to make transfer possible, even across scripts, %thereby reinforcing the findings of \citet{pires2019multilingual} on Hindi and Urdu, 
    in a scenario where the target language is not in the pre-training corpora.
    %\item \draftHL{We present detailed analysis}{C'est pas ca qui confirme le point 1 ?} demonstrating that \mbert is actually making use of its multilingual representations to process \arz.
    
    %that  showing how code-mixing impacts the modelling ability of multilingual language models in these different scenarios.
\end{itemize}

%Based on this, we study the following scenarios (shedding light on questions (1) and (2)). 

% - can it model : 1- MLMacuracy   1- can it detect ? 2- can it transfer ? 
% - can it learn : can it learn in low rssource ? How many sentences ? can we adapt it ? 

%it is still unclear how much such a model can capture a 
%we will shed light to the following question. Is it possible for a given multilingual pretrained model to transfer to a new 

% sucess of contextual embedding... 

% mBERT

%Moreover, to this day, arabizi does not appear in any released pretrained contextual models. Particularly, mBERT whose been trained on 104 languages that includes French, Arabic but not arabizi. 

% COMPARE ARABIZI to ARABIC and FRENCH ?? 

%For those reasons, Arabizi is a tough challenge for Natural Language Processing. More specifically, Arabizi gives us a great opportunity in analysing the behavior of mBERT on such different data. 

%We will start by describing some core properties of Arabizi. We will then present our results on Part-of-Speech Tagging and Dependency Parsing on 4 different scenarios. In a nutshell, we show that mBERT leads to similar performances compared to a non-contextual strong baseline... 

%\input{djame_introduction_dontouch}

\section{Related Work}
\label{sec:related}
% Ok je suis passé dessus

% https://www.aclweb.org/anthology/D19-6106.pdf
% How multilingual is multilingual BERT ? 
% XLM ? 
% Low ressource Transfer Soogaard ? 
% 
\paragraph{Word embedding \& Cross lingual Transfer}

%Research on cross lingual transfer has a long tradition. % In the recent years, it has largely benefited from the successes of word embedding \cite{mikolov2013distributed,pennington2014glove,mikolov2013exploiting,smith2017offline} %efficiently merge several monolingual spaces into a single multilingual one with limited supervision. 
%Based on those common representations \cite{guo-etal-2015-cross, } efficiently transferred from one source language to a target one. 
%\paragraph{Cross lingual Transfer}
Recently, cross lingual transfer has benefited from multilingual language models. We refer to \cite{lample2019cross,eisenschlos2019multifit,vania-etal-2019-systematic,wu2019emerging,conneau2019unsupervised,wu-dredze-2019-beto} who demonstrate the efficiency of language models in zero-shot transfer settings for a variety of tasks. In this regard, \citet{pires2019multilingual} analyze in detail the zero-shot transfer ability of \mbert on sequence labeling. \citet{wang2019cross} suggest that cross-lingual transfer of multilingual models rely on \textit{structural properties} of languages. Both studies focus on transfer between languages that are part of the pretraining corpora. In our work, we study the ability of \mbert %as a multilingual model 
to transfer to an unseen language. %We use POS tagging and dependency parsing as our case studies.

\paragraph{Code-Switching}is a hard challenge for NLP as shown in the myriad of works that have tackled this phenomenon for more than 10 years, see for example \cite{solorio-liu-2008-part,vyas-etal-2014-pos,cetinoglu-coltekin-2016-part,lynn-scannell-2019-code}.
%%% cite Solorio and Liu (2008), Vyas et al. (2014), Cetinoglu et al.(2016)
\newcite{ball-garrette-2018-part} and \newcite{pires2019multilingual} analyzed the performance of neural models for sequence labeling showing that those approaches can cope with such a complexity.
In our work, we face both code-switched and highly variable data.%, making \arz a great and extreme \draftreplace{study case}{case study}.\draftnote{super redondant avec la phrase finale de la section}

% TALK ABOUT kondratyuk201975

%The generalization ability across languages of multilingual models has been extensively studied for languages that appear in the training corpora. 

%\paragraph{Natural Language Processing on Arabic dialects}
%Research on Arabic dialects is quite extensive. Space is lacking to describe it exhaustively. In relation to our work, we refer to 
%\cite{samih2017neural,zalmout2019joint} that recently efficiently adapted neural architecture to perform word segmentation, lemmatisation morphological analysis and POS tagging on an Arabic dialect. %\cite{cotterell2014algerian} 
%To the best of our knowledge, our work is the first to analyse the ability of multilingual language models on an Arabic dialect.%\bm{add Cotterel}

\paragraph{Unsupervised Adaptation of Language Models} 
%if used afterwork  
%To make use of the English version of BERT on English tweets and old English data, 
\newcite{han2019unsupervised} show that fine-tuning BERT in an unsupervised way using its masked language objective brings significant improvement to downstream sequence labeling tasks for out-of-domain Old English. Studying the specific case of English-Spanish code-mixing, \newcite{gonen2018language} show how to adapt bilingual language models to code-mixed data. 
In our work, we focus on unsupervised adaptation and analyze its impact on the even more challenging case of \arz. %\Djame{redondant avec la phrase que j'ai pointé plus haut}

%%%% section part

% DS: ok j'ai fini là

\section{\arz}
\label{sec:arabizi}
%\bm{add more ref to Arz + maltese parent}\\
Arabic varieties are  often classified into three categories \cite{habash10intro}: (i) Classical Arabic, as found in the {\em Qur'an} and related canonical texts, (ii) Modern Standard Arabic (MSA), the official language of the vast
majority of Arabic speaking countries and (iii) Dialectal Arabic. %whose instances exhibits so much variations that they are not understandable across geographically-distant regions. 
This work focuses on North-African dialectal Arabic in its Algerian form, understood and spoken  by more than 40 million people in the Maghreb \cite{magreblanguages}. In its written form, it is mostly found online and in Latin script. For simplicity we refer to this North-African  Arabic dialect as North-African Arabizi \cite{farrag2012arabizi} or \arz, illustrated here:%\ref{ex:un}.}
\\[1mm]
%\begin{center}
\indent\begin{tabular}{l}
source: {\tt Mrhba, Ana 3rbi mn dzaye}\\
%\end{center}
translation:{\textit{``Hey, I’m Arab from Algeria''}}
\end{tabular}
\\[1mm]
%\end{center}
%\vspace{-0.2cm}
%\enumsentence{\label{ex:un}
%\begin{center}
%}
\indent Like other written languages found on social media and even more importantly as it is not standardized,\footnote{I.e.~no writing rules are officially defined.} \arz shows a high degree of variability across writers. As part of its variability, %\draftreplace{\arz is code-mixed with French.}{
\arz frequently involves code-switching with French.%}%}, especially with French for historical reasons.}

%We list the inherent key properties of \arz. First, Unlike MSA. diacritical signs are replaced  by  actual vowels.  Second, writers make use of digit numbers to cope with Arabic letters absent from the ASCII\footnote{https://www.ascii-code.com/} character sets. Third, there is a high degree of variability between \arz writers as no standard writing rules have been defined. Finally, It shows a high degree of code-mixing with French.

%\begin{itemize}
%\item
%\item Use of digit numbers to cope with Arabic letters absent from the ASCII character set
%\item High degree of variability between \arz writers. 
%\item High degree of code-mixing with French
%end{itemize}  

Moreover, \arz does not belong to the pre-training corpora of \mbert. For this reason, we take \arz as our case study to analyze the ability of \mbert to handle an unseen, highly variable and code-mixed dialect. 

\paragraph{Data}

The data we use comes from two main sources. The first one, described by \newcite{cotterell2014algerian}, is a collection of 9000 raw Algerian romanized Arabic sentences, a sample of which  has  been annotated with Universal Dependency trees \cite{mcdonald-etal-2013-universal} and word-level language identification\footnote{\arz and French prevalence in the train set (\% token): \arz $64.64\%$, French $33.84\%$, then MSA, English \& Spanish.}  by \newcite{seddah2020} totalling 1,434 (1172/146/178) annotated sentences. Our second source, also released by \newcite{seddah2020}, is a collection of 49,546 raw \arz sentences. %In all scenarios, we do not make use of annotated \arz data for training. %Therefore, to better support our conclusions, we use the train set (1,172 sentences) for evaluation. 

%\begin{table}[h!]
%\small
%\centering
%\begin{tabular}{rll}
%\toprule
%  Data & Sentences & Tokens\\
%  \midrule
%  Raw data   & 49,546 & 1,586,538\\
%   POS Annotated   & 1,434 & 22,465\\
%  
%\bottomrule
%\end{tabular}
%\caption{Summary of \arz data. The raw dataset is the concatanation of the raw data found in \cite{cotterell2014algerian} and the one found in \cite{srivastava:hal-02270527} (In annotated, 1,172 sentences for
%  training, 146 for validation and 148 for test)}
%\label{tab:stats}
%\end{table}
%\label{data}

%The POS dataset is composed of Universal Dependancies (UD) \cite{petrov2011universal}  POS tags-sets as well as token-level language identifiers among \{\arz, French, , MSA, Spanish, English\}, with a strong prevalence for \arz and French.% We refer to \cite{Srivastava} for an extensive description of the data. 

\paragraph{Baselines\label{sec:baselines}} To grasp the complexity of \arz, we run some preliminary experiments. We take \newcite{qi2019universal}'s tagger and parser as our strong baselines % (\stanford). %\cite{straka-etal-2016-udpipe} (referred as \udpipe),  \stanford as a neural model \cite{qi2019universal} 
(named \stanford\footnote{Ranked top 3 (after correction) in POS tagging and parsing at the 2018 UD shared task, trained using French fastText vectors \cite{mikolov2018advances}.}), and as our bottom lines the majority class predictor for POS tagging and the \textit{left predictor}\footnote{Whereby each word is attached to its immediate left neighbor.} for dependency parsing. %\footnote{more precisely : identifies punctuation tokens and otherwise predict the majority class (the NOUN class)}
%These results show the complexity of processing \arz data. 
Competitive taggers perform on datasets of similar size above 90\%. \stanford only reaches 84.20\% on our data for POS tagging and 52.84\% for parsing, as measured by the unlabeled attachment score (UAS; cf. Table  \ref{tab:all_details}). 
% baselines for all our experiments.

%\bm{present data for other languages we use}
%\paragraph{Challenge for mBERT}
% OOD
% How much cut rate ??
% 

%\begin{table}[b]
%\centering\small
%\begin{tabular}{lc}
%    \toprule
%    Model & Accuracy \\
%    \midrule
%    \dummy &  20.49 \\
%    \udpipe &  73.10\\
%    \stanford & 84.20  \\
%    \bottomrule
% --> lang detect 10128071 

%\end{tabular}
%\caption{Preliminary experiments : reporting POS accuracy on %model trained on \arz treebank (training (resp. validation) on 1,172 (resp. 146 sentences), test on 148 sentences \label{baselines2}
%}

%\end{table}

\section{Model}

%Our work focuses on the transfer abilities of the multilingual version of BERT. 
\mbert is a Transformer \cite{vaswani2017attention} trained as a joint masked-language and a next sentence prediction model on sub-word level tokenized sentences. More details can be found in \cite{devlin2018bert}. We use the multilingual cased version of BERT. %\footnote{https://github.com/google-research/bert/blob/master/multilingual.md}. 
\mbert was trained on the concatenation of the Wikipedia corpora for 104 languages. %In regard to our work, we emphasize that French and Arabic (MSA) are part of the the training corpora while Maltese and \arz are not. %As input, BERT uses wordpiece tokens introduced in \cite{schuster2012japanese} and named as such in \cite{wu2016google}. In a nutshell, wordpiece tokenization splits tokens into sub-words based on their frequency in the pre-training corpora. 
\paragraph{POS tagging and dependency parsing with \mbert}
Following \newcite{devlin2018bert}, we turn \mbert into a POS tagger by appending a softmax on top of its last layer. For parsing, we append the biaffine graph parser layers described by  \citet{dozat2016deep}. In both cases, we fine-tune the overall model by backpropagating only through the first sub-word token of each word. %POS tagging is a token level task while BERT tokenizes token into frequent sub-tokens. To predict and train on tags, we only backpropagate on sub-tokens that are a the beginning of each token. 
%By \textit{supervised fine-tuning} we refer to the process of training \mbert on POS tagging. 
We call these architectures \mbertpos and \mbertparse, and use them in our zero-shot learning experiments by applying models trained on a source language to data in our target language.
%\bm{add parsing}

\paragraph{Unsupervised Adaptation}

We call \textit{unsupervised adaptation} the process of fine-tuning \mbert in an unsupervised manner using its Masked-Language Model (MLM) objective trained on raw sentences. We refer to \mbert fine-tuned on raw data as  \mbertadapt. We define as \mbertadapttask, with \textsc{Task} referring to \textsc{Pos}, resp. \textsc{Parse}, to point to \mbertadapt fine-tuned as a POS tagger, resp. parser. 

\section{Experiments}
%\mbert is pre-trained on 104 languages. 
Our goal is to measure how well \mbert makes use of its multilingual pre-training on an unseen dialect. %Using POS tagging and parsing as our case studies, we focus 
%on cross-lingual transfer in which no \arz annotated data is seen . 
We defined a \textit{source} language as a language on which a POS tagger or a parser are trained, and will report the performance of the resulting models when applied to \arz data.% (our target). %: zero-shot cross-lingual transfer and  unsupervised adapted cross-lingual transfer. %and low resource.

\iffalse
\subsection{Scenarios}
\paragraph{Zero-shot Cross-lingual Transfer}
In this scenario, we train \mberttask on a source language and evaluate on \arz. Here, no \arz data is seen by the model at any stage of the training process. We measure how well it transfers the representations learnt during pre-training on 104 languages and during fine-tuning on the specific task on a source language to process \arz.

\paragraph{Unsupervised adapted cross-lingual transfer}
In this second scenario, we measure how making use of raw \arz sentences impacts downstream cross-lingual POS tagging and paring. To do so, first we fine-tune \mbert in an unsupervised way using its masked language model  (unsupervised adaptation). Then, we run supervised fine-tuning on a source language and evaluate it on \arz.
In this scenario, \mbert only adapts to \arz with 49,546 raw sentences but no annotated data.  
\fi 
\subsection{Source Languages}
\label{sec:sourcelanguages}
We study transfer along two independent directions. The first one is the relatedness of the source language to \arz. The more different they are, the worse we expect the transfer to be. Our second direction distinguishes between source languages included in \mbert pre-training corpora and those that are not. We expect the transfer to be better when the source language is included in the pre-training corpora. To cover the full scope of cases, we pick Modern Standard Arabic, French\iffalse, Italian\fi{}, English and Vietnamese. 

As recalled in \cite{habash10intro,vceplo2016mutual},  Maltese is related to the Arabic continuum of languages. It is standardized and written in an extended Latin script. This makes Maltese a promising candidate for transferring to \arz. 

We also use French in order to study the impact of code-mixing. In addition, we experiment with English as another European language written in Latin script, but which is not code-mixed with \arz. Finally, we use Vietnamese as the most unrelated language to test the cross-lingual power of the model in the most extreme case.  
We refer the reader to the Appendix (Table~\ref{tab:languages}) for an overview of the source languages. 
We sample the training datasets to have  1,200 sentences for each source language.\footnote{We pick the first 1,200 training sentences. More information on the datasets used is given in Appendix~\ref{sec:sourcelanguages}.} Additionally, we report results in the standard supervised setting in which we fine-tune \mberttask on \arz and evaluate on \arz. This provides us with an upper bound on how we can expect \mbert to perform on such a language.

%\iffalse % BENJAMIN: met ça dans l'annexe.

%\fi

%We present experiments run on French, Maltese, Modern Standard Arabic and Vietnamese. Our goal is to understand the transfer ability of \mbert in POS tagging in regard to the source languages. French and MSA are the two closest languages (French due to the code-mixing and MSA because \arz belongs to the family of Arabic dialects although not sharing the same script). Maltese is descended \cite{malta} from an Arabic dialect and is written is Latin script. We control for the amount of training instances and pick treebanks with around 1,500 sentences in the train set\footnote{can be found in the Appendix \ref{?} for more information on the data we use.} 

%\paragraph{Languages}
%\paragraph{Protocol}

\subsection{Optimisation}
For supervised fine-tuning, we use the same range of hyper-parameters as \newcite{devlin2018bert}. For unsupervised adaptation, we run preliminary experiments to measure the impact of the raw corpus among the 49,000 sentences \arz corpus, and \arz mixed with a sub-sample extracted from \mbert pre-training corpora. As reported by \newcite{gonen2018language}, we found that, if carefully optimized, fine-tuning \mbert with its masked language objective directly on the target data leads to the best models.\footnote{Cf.~Appendix \S~\ref{hyper} for details on hyper-parameters.}

%\subsection{Comparison}
%We compare to strong baselines including \stanford. Moreover, we expect the transfer of \mbert to happen thanks to its multilingual pre-training. To study this specific point we 
%we specifically study how it behaves in regard to the code-mixing degree, one of the core property of \arz\\

%To enrich our analysis, for each of these scenarios, we also compare \mbert to a randomly initialized transformer trained as a masked-language model on the 50,000 \arz sentences (hereafter \mlm and \mlmpos when fine-tuned on POS). This comparison shows the impact of the pre-training on 104 languages when transferred to \arz compared to a \arz specific language model

% ? what to say here 

% bert briefly 
% mBERT , pretrained on what 
% p() transfer learning
% task-specific supervised finetuning 
% unsupervised fine tuning 

\section{Results and Discussion}
\label{result}

We present our results in Table~\ref{tab:all_details}. We report the accuracy for POS tagging and the Unlabeled Attachment Score (UAS) for parsing.\footnote{For parsing, we focus on UAS only and do not report  Labeled Attachment Scores (LAS). We do so because we noticed
diverging conventions in the labeling scheme between the \arz treebank and the other treebanks we use, which result from different annotation choices
allowed by the Universal Dependencies framework.} Any performance above the bottom line demonstrates that transfer is happening from the pre-training or fine-tuning stages to process \arz.  For both POS tagging and parsing, \mberttask performs outperforms the baselines by a large margin when the source is Maltese, French and English. This shows that \mbert is able to transfer to \arz even without having been trained on any \arz tokens at any stage of the training process. %French language leads to strong performance on target \arz for POS tagging (resp. parsing) with 33.12 (resp. 38.54)\% accuracy in zero-shot transfer and 47.32 (resp. 43.77) \% after unsupervised adaptation. We make the hypothesis that these performances are due to the high-code mixing proportion.

%In both scenarios, French is the source language that leads to the best performance on target \arz with 41.81\% accuracy in zero-shot transfer and 50.99\% after unsupervised adaptation. We make the hypothesis that these performances are due to the high-code mixing proportion.

% BOOST :POS (38.94+47.32+45.59+28.08+23.)-(35.13+33.12+30.67+16.55+16.92) / 5
% UAS : (40.04+38.54+32.40+28.23+13.98)-(42.32+43.77+38.83+34.34+14.44)

We report an average boost with \mbertadapttask of +10.15 points in POS tagging and +4.10 in UAS across all source languages when compared with \mberttask. This means that the unsupervised adaptation on 49,546 raw \arz sentences is efficient even on such an out-of-domain language. In all these settings, \stanford, the very strong neural baseline, designed specifically for POS tagging and parsing, is outperformed by \mbert (it only reaches 31.90 and 33.74 (resp. 15.53) for parsing in UAS (resp. in LAS) on \arz  when trained on French)%\bm{should I have all the scores in appendix?}. 

\subsection{Cross-script Transfer}
In zero-shot settings, cross-script transfer does not perform above the bottom lines when the source is MSA in POS tagging. %No transfer happens between languages that do not share the same script whether it is MSA or Vietnamese. 
%This result contrasts with what is observed in \cite{pires2019multilingual} for Urdu and Hindi. 
Our hypothesis is that such transfer requires the target language to be in the pre-training corpora as reported by \newcite{pires2019multilingual} in the case of Urdu and Hindi. Nevertheless, to our surprise, we observe an impressive +13 boost in tagging and +6.11 in parsing performance after unsupervised adaptation when the source is MSA,  outperforming the baselines. This means that in the case of MSA, cross-script transfer happens when the target language is seen during unsupervised adaptation, and 49k sentences are enough to lead to such a transfer. Moreover, cross-script transfer is better with MSA than Vietnamese, suggesting that the multilingual model is making use of the proximity of MSA and \arz. 

%\paragraph{Cross-script transfer}
%First, \mbert is not able to transfer from MSA to \arz. Indeed, the model trained on MSA is bellow the \dummy baseline. We recall that MSA is written in Arabic script while \arz in Latin. This leads to a vocabulary total mismatch. This result contrasts with what is observed in \cite{pires2019multilingual} for Urdu and Hindi. \mbert do not transfer across scripts in the case of unseen \arz. The performance is similar to \mbertpos tuned on Vietnamese showing that \mbert is not able to capture any  of the common roots of Arabic and \arz. Our explanation is that cross-script transfer requires the source and the target to be in the pre-training corpora (as Urdu and Hindi are).  

\begin{figure}[t]
\centering
\includegraphics[width=8.0cm]{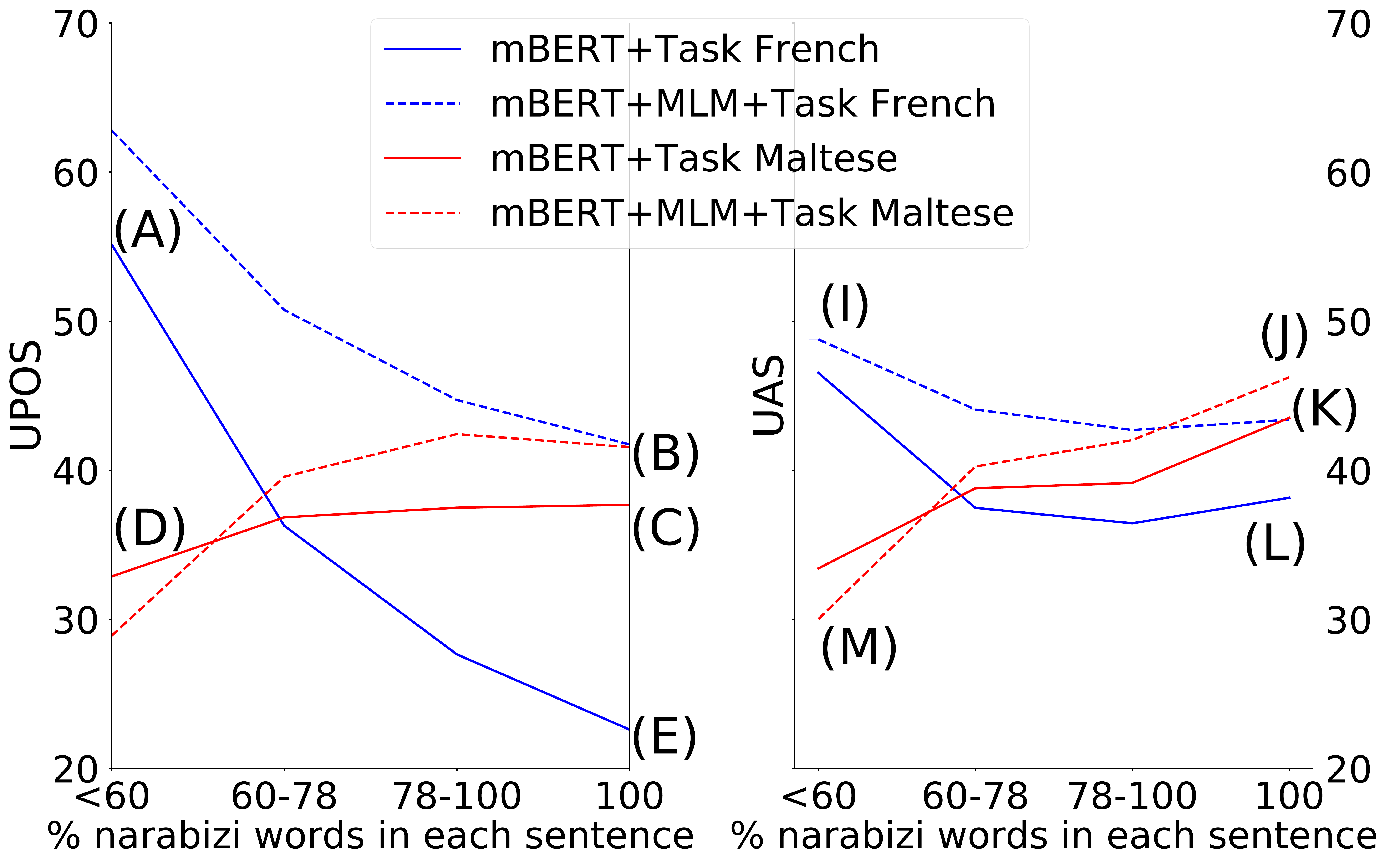}

\caption{Performance with regard to code-mixing rate, reported on \arz train set to have enough data per bucket.  (5 seeds). (X) markers commented in sec.~\ref{sec:code_mix}. NB: no \arz annotated training data seen during fine-tuning.}
\label{fig:code_mixe_curve}
\end{figure}
%\bm{should have parsing also}
\subsection{Impact of code-mixing}
\label{sec:code_mix}
We hypothesize that the high level of transfer when the source is French is due to the high code-mixing proportion of \arz. 
To test our hypothesis,
%Moreover, \mbertpos tuned on French outperforms all compared models significantly, including \stanford (+7.21). To understand how the code-mixing is at play
we present in Figure \ref{fig:code_mixe_curve} the performance of the model with respect to the code-mixing ratio. We split the dataset into four buckets of around 25\% of the full dataset, according to the ratio of native \arz tokens in each sentence (between less than 60\% to stricly 100\%) as opposed to French tokens. %We get 4 buckets (from right to left in Fig.\ref{fig:code_mixe_curve}):  100\%,  between 78 and 100\%, between 60 and 78\%  below 60\% of percentage of \arz token in each sentence.  
We compare French and Maltese as source languages. We confirm our intuition that code-mixing explains the good performance of the model trained on French. Indeed, on sentences that have 100\% \arz tokens, \mberttask trained on French performs poorly (cf. fig.~\ref{fig:code_mixe_curve} (E) for POS and (L) for parsing). On the other side, for sentences that include at least 40\% of French tokens, scores reach 54\%  (cf. (A)) for POS tagging and 47\% for parsing (cf. (I)).
Moreover, for French, \mbertadapttask leads to an impressive 21.2\% error reduction compared to \mberttask  for POS tagging (33.12 vs.~47.32) and an 8.5\% error reduction for parsing (cf.~Table~\ref{tab:all_details}). We observe in Fig.~\ref{fig:code_mixe_curve} (cf.~(B) and (K)) that this improvement mostly comes from a better accuracy on \arz tokens. Interestingly, we observe that unsupervised fine-tuning leads to the closing of the gap between the performance of the models tuned on French and Maltese on native \arz tokens (+15: (B)-(E)  vs.~+2.4: (B)-(C) for POS tagging and +5.6: (K)-(L) vs. +2.2: (J)-(K) for parsing). This demonstrates the capacity of unsupervised fine-tuning to close lexical mismatch between distant languages such as native \arz and French. %To a smaller extent, we also observe this behavior for parsing (cf. (J) and (I)).
%by comparing the difference of the gap before and after unsupervsied fine-tuning 
%the cross lingual performance between French and Maltese, 
%we observe that while \mbertpos tuned on Maltese performs much better on native \arz sentences than \mbertpos tuned on French (+15 points cf. (A) vs (E)), \mbertadaptpos tuned on Maltese performs nearly as well as \mbertadaptpos tuned on French (cf. (C) vs (G)). This shows the power of unsupervised fine-tuning. %It leads to closing the gap between vocabulary and structural gaps.

%By comparing this result with \stanford, we observe that the main difference come from the performance on code-mixed text between 60\% and 78\% of French. %Despite not having seen \arz token at train time, \mbert is much more robust than the state-of-the-art model on such code-mixed data.  %Indeed, it outperforms \stanford  by 6 points \bm{add score stanford in plot}.
%\paragraph{Impact of \mbert pre-training}
%On French \mbertpos performs 3 points above \mlmpos (cf. Table \ref{tab:all_details} (i) and (iv), 39.11 vs 36.23). This result quantifies the transfer happening between the pre-training stage and the downstream POS tagging. \mbert is able to make use of what is learnt during the pre-training stage for \arz POS tagging. 
\subsection{Transfer between unseen languages}
Surprisingly, \mberttask tuned on Maltese does not perform poorly. It leads to the best performance for \mberttask for both POS tagging and parsing. It outperforms \stanford in the zero-shot scenario by 5 points in POS tagging and 6 points in parsing. As seen in Figure~\ref{fig:code_mixe_curve} (C) and (J), it performs the best on native \arz sentences (with no code-mixing). This result is surprising as Maltese is absent from the pre-training corpora. It shows that \mbert is able to capture \textit{structural properties} shared by related languages even if they are absent from the pre-training corpora, thereby extending the observations described by \citet{wang2019cross}.  

\begin{table}[t]
\resizebox{0.97\linewidth}{!}{ % was .7775
\footnotesize\centering
\begin{tabular}{lrrrr}
    \toprule
    %Scenario  & 
     % Source & \mbertpos  & \mbertadaptpos & $\cap$ Vocab\\
      & \multicolumn{2}{c}{\underline{\mberttask}}  &\multicolumn{2}{r}{\underline{\mbertadapttask}} \\%& $\cap$ Vocab\\
%            &            & +Task \\%& \\
            %& UPOS / LAS  & UPOS / LAS    
        % &\cline{2-3}&\cline{4-5} %& %\cmidrule{4-5} 
        \textit{Source}    & POS & UAS  & ~~~~~~~~~~~~~~~~~~POS & UAS    \\    % 
    \midrule    
    %\multicolumn{1}{c}{(i) \textit{zero-shot}} & \multicolumn{2}{c}{}\\
    %Zero shot Cross-lingual transfer & mBERT & fr\_gsd && 35.57  \\ 
    %Zero shot transfer & 
    %French partut & \textbf{ 41.81} & 50.99\\% & 32.63\\
    %with LAS : French (GSD) & \textbf{33.12~/~16.86}   & 47.32 / 24.01 \\
    Maltese & \textbf{35.13} & \textbf{40.04} & 38.94 & 42.32\\
    French & 33.12 & 38.54  & \textbf{47.32} & \textbf{43.77} \\% UAS % (LAS : 17.34 10491000)
%    Italian partut & 36.78 &  44.3 2  & 38.66\\ % MLM=tr:47.07
    %English partut & 30.67 & 39.61 \\ %& 33.81\\%
    %  LAS English & 30.67 / 13.47 & 45.59 / 19.80 \\ %& 33.81\\%MLMtr:41.80
   English & 30.67 & 32.40 & 44.59 & 38.83 \\ %& 33.81\\ UAS 
    % LAS Maltese & 35.13 / 17.81 & 38.94 / 18.10
    
    % 10491000 LAS: 18.29 
    % & 35.13 \\  %MLM tr:46.11
    % LAS MS Arabic & 16.55 / 9.27 & 28.08 /  \\% & 0\\  % MLM:(tr:29.14)

    MSA & 16.55 & 28.23 & 28.08 & 34.34  \\% & 0\\  % MLM:(tr:29.14)
    %Chinese & 6.53 &  8.84 \\ %& 0 \\ % tr:24.62
    Viet. & 16.92 & 13.98 &  23.21 & 14.44 \\ % tr:24.62
    \cdashline{1-5}
    % LAS 10491000 \arz       &  /56.60      &   /    \\% 
   \arz   &   81.60 & 66.84      &   82.61 & 67.12    \\% job POS 10491294
    % LAS \mbertadapt 47.71 job : 
    \bottomrule
    & \multicolumn{4}{c}{\textit{Baselines}}\\ 

    %\multicolumn{1}{c}{} & \multicolumn{1}{r}{\stanford} & \multicolumn{1}{r}{\dummy}\\
   & \multicolumn{2}{c}{\underline{~\stanford~}} & \multicolumn{2}{c}{~~~~~~~~~~~~~~~~~\underline{~~~~Bottom  lines~~~~}}\\
     %Bottom  line \S3\\
    %\multicolumn{2}{r}{\dummy}\\
    %Zero shot transfer & 
   % Random+POS & French  & 30.29 \\
    %Zero shot transfer &
    \arz    & 84.20 & 52.84 & 20.49 & 18.71\\
     French & 27.00 & 33.74 &  -    & - \\ % LAS is 16.14
    %En & 17.79 / 28.11 &  \\ %EWT LAS is 4.5
    % French  & 31.90 /  & - \\
    % French*  & 32.50 / & - \\
%     \textcolor{red}{\bf !!French**!}  & \textcolor{red}{\bf 32.50} & - \\

    \bottomrule
\end{tabular}
}%end of scalebox
\caption{Cross-Lingual performance averaged on 5 seeds on the \arz test set. Baselines are described in Section~\ref{sec:arabizi}.}
\label{tab:all_details}

 %``$\cap$~Vocab'' is the percentage of tokens found simultaneously in source language, \arz (train) and \mbert vocabularies. 
%\\ * using \arz fastText vectors (trained on 50k sent)\\

%\bm{** case to show that even with 50k sentences we beat stanfordnlp}
%\Djame{Je suis pas sur qu'utiliser des embeddings narabizi pour un truc entrainé sur le français soit super informatif. Je virerais ça}}

\end{table}

%EXPE TO RUN : Our hypothesis is the following. mBERT uses positional information to predict unseen word forms. To prove our point we train on the exact same setting, but we shift the word embedding index. This is equivalent to make mBERT blind to word-piece information. We report the same experiments on Table XX. 
\paragraph{Is the multilingualism of \mbert at play?}
Finally, we want to show that the ability of \mbert to achieve cross-lingual transfer is related to the 104 languages it is pre-trained on, rather than because a pre-trained Transformer is an inherently good POS tagger or parser. To do so, we compare \mbert with three other models: Roberta, the optimized English version of BERT \cite{liu2019roberta}, \camembert (C.BERT) the French version of BERT \cite{2019arXiv191103894M}, and a randomly initialized \mbert-like Transformer (\random). %(to show the capacity of the arhitecture to perform POS tagging).  %and \mbert after randomly shuffling its wordpieces vocabulary (\mbertshuffled).
%\random shows the capacity of the architectur to perform POS tagging as such.%impact of pre-training. % while comparing with \mbertshuffled gives us the transfer coming from the specific wordpieces\bm{wordpieces should be mentioned before}. 
We focus our analysis on French and Maltese.  \mbert is the model that leads to the most successful transfer in both cases and for both tasks, by a very large margin in the case of Maltese. This shows that pre-training on such a diversity of languages is at the core of the transfer to \arz. 
%Surprisingly, BERT and CamemBERT do not perform poorly in most cases showing that monolingual models can transfer from unseen to unseen languages also. CamemBERT is particularly accurate. Our hypothesis is that the scale of data used for pre-training is at play (10x more data used for CamemBERT than the two others). We will let this for future research. In the case of Maltese, \mbert outperforms all other version significantly. We interpret this result as another clue highlighting the capacity of a multilingual model to transfer its multilingual representation to unseen languages.  %As presented in table \ref{tab:berts} we show that \mbert outperforms all other versions of BERT when the source language is not the pre-training language. 
%e make therefore emphasize the following question : How does multingual BERT pre-trained on 104 languages including French is able to correctly predict POS tags on unknown and unseed Arabizi ? 

\begin{table}[h!]
\footnotesize
\centering
\scalebox{.9}{
\begin{tabular}{l@{\hspace*{3mm}}c@{\hspace*{1.2mm}}c@{\hspace*{3mm}} c@{\hspace*{1.2mm}}c@{\hspace*{3mm}}c @{\hspace*{1.2mm}}c@{\hspace*{3mm}}c@{\hspace*{1.2mm}}c}
\toprule
    % & \mbert & R.BERTa & C.BERT & Rand.\\ %& MLM \\%& \mbertshuffled \\
    & \multicolumn{2}{c}{\underline{~~~\mbert~~~}}& \multicolumn{2}{c}{\underline{RoBERTA}}& \multicolumn{2}{c}{\underline{~~C.BERT~~}}& \multicolumn{2}{c}{\underline{~~~~Rand.~~~~}}\\
       & POS & UAS  & ~~~POS & UAS & ~~~~POS & UAS & POS & UAS \\%& /36.91\\%& 29.78  \\ BERT : 10483155 pos job;  RANDOM job optimize : 10491315 ;  10483156 parisng job  Parsing results : 10482946 job POS 10482961 job

  \midrule
  %\arz &  & - &  \\
  
  fr & \textbf{33.12}&\textbf{38.54} & 29.75&27.41 & 31.77&32.55 & 30.29&25.30 \\%& /36.91\\%& 29.78  \\ BERT : 10483155 pos job;  RANDOM job optimize : 10491315 ;  10483156 parisng job  Parsing results : 10482946 job POS 10482961 job
  %French & \textbf{41.81} & 37.37 & 40.37/32.55 & 30.29 \\%& 29.78  \\
  %Italian   & 39.71 & 33.75  & 41.21 \\
  %English   & 37.67  & 27.75 & 40.30 \\
  %Maltese   & \textbf{35.13} & 28.38 & 22.82 & 19.81 \\%& 19.32 \\
  mt   & \textbf{35.13}&\textbf{40.04} & 25.45&17.27 & 31.62&34.65 & 19.81&19.04 \\ %& / 28.72 \\%& 19.32 \\
 %En & 30.67/32.40 & 17.01/22.39  & ?/30.46 &  \\% JOB camembert from english : 10492188 parsing , 10492189 POS . 
  % JOB 10492111 roberta POS tagging ; parsing job 10492112 parsing ; NEW PARSING RANDOM OPTIMIZED : 10492066 
  %Arabic (MSA)   &   &  &  \\
  %Vietnamese   &  &  &  \\
  
%LAS : score MLM model : 10490988 job
%French GSD && 17.79(0.55)(20.00) \\ 
%Maltese && 7.41(0.47)(20.00) \\ 
%narabizi-ud-train && 47.32(0.09)(20.00) 

\bottomrule
\end{tabular}
}
\caption[]{Zero-shot transfer from French (fr) and Maltese (mt) to \arz. 5 averaged seeds.}

\label{tab:berts}
\end{table}

\vspace{-0.2cm}
\section{Conclusion}
Our work on \arz reveals novel properties of multilingual language models.  %language models in their abilities to process and learn unseen languages.
We have shown that transfer learning approaches can be used successfully on this language, both in zero-shot scenarios, where no target language data is used at any stage of the training process, and in unsupervised adaptation scenarios, where only raw target language data is used.

This is remarkable, because \arz, an increasingly used language on social media, is an extremely challenging language for NLP in general and transfer learning approaches in particular, for at least three reasons: (i)~it is written in a script different to its closest resourced relative (Modern Standard Arabic), (ii)~it displays a high degree of variation because of the lack of spelling standard, and (iii)~it involves frequent code-switching with an unrelated language (French, in our case).

Our results pave the way to using transfer learning approaches to build NLP tools not only for \arz but also for other vernacular varieties of Arabic written in Latin script, and more generally for any low resource language, even when it displays some of the challenging properties listed above. Our paper therefore sheds light on a way to initiate the development of NLP ecosystems for languages and language varieties that are increasingly used online, for which NLP  is badly needed, but for which few resources, if any, are available to date.

%by making use of related languages in transferring to unseen dialect even when the source is also unseen or .%. Moreover, we prove that Unsupervised Adaptation makes cross-script transfer possible. % languages astransferring across script and even when the source language are unseen. 

%Cross-lingual transfer abilities of multilingual pretrained language models have been extensively studied in the literature for a great variety of tasks and languages.  Experimenting on \arz, a highly variable and code-mixed, reveals that transfer learning using multilingual language models can be pushed to the extreme of an unseen dialect. These results stand as a new clue toward the interlingual representation of multilingual models. % and that it highly benefits from unsupervised adaptation. 
\section*{Acknowledgments}
We want to thanks Yanai Elazar, Ganesh Jawahar and Louis Martin for  proofreading and insightful comments.
This work was partly funded by two French National funded projects granted to Inria and other partners by the Agence Nationale de la Recherche, namely projects PARSITI (ANR-16-CE33-0021) and SoSweet (ANR-15-CE38-0011), as well as by the second author's chair in the PRAIRIE institute funded by the French national agency ANR as part of the ``Investissements d’avenir'' programme under the reference \mbox{ANR-19-P3IA-0001}. This project also received support from the French Ministry of Industry and Ministry of Foreign Affairs via the PHC Maimonide France-Israel cooperation programme.

%\textit{It can perform transfer to a certain extent between very different languages even on strict arabizi : It is highly adaptable. Indeed, unsupervised adaption is an efficient strategy, while supervised fine-tuning.  
%- Maltese for interlingua
%This evidence suggest the possibility of an interlingua that can easily transfer from unseen }

%\noindent \textbf{Preparing References:} \\
%Include your own bib file like this:
%\verb|\bibliographystyle{acl_natbib}|
%\verb|\bibliography{acl2019}| 

%where \verb|acl2019| corresponds to a acl2019.bib file.
\bibliographystyle{acl_natbib}
\bibliography{acl2019}

% APPENDIX

%\newpage
%.
\newpage
%\onecolumn 

\appendix

\section{Appendix}

\subsection{Source Languages}

\begin{table}[h!]
\small
\centering
\begin{tabular}{rlll}
\toprule
   Language & script & relatedness & $\in \Omega\textsubscript{\mbert}$ \\
  \midrule
  
  \arz & Latin & - & no \\
  French & Latin  & code-mixed & yes \\
%  Italian   & Latin & none & yes \\
  English   & Latin & none & yes \\
  
  Maltese   & Latin  & shared root & no \\
  MS Arabic   &  Arabic & shared root & yes \\
  Vietnamese   & Latin & none & yes \\
  %Chinese   & Chinese & none & yes \\
  %% MSA est defini en haut man,c std.
\bottomrule
\end{tabular}
\caption[]{Source language in regard to \arz based on languages relatedness and inclusion in model pre-training corpora ($\in \Omega\textsubscript{\mbert}$ for languages included in the 104 pre-training languages of \mbert).}

\label{tab:languages}
\end{table}

\begin{table}[h!]
\footnotesize
\centering
\begin{tabular}{lrr}
\toprule
%   Language  & dataset    \\%& \mbertshuffled \\
%  \midrule
  %\arz &  & - &  \\
  %French & fr\_partut \\%& 29.78  \\
  French & fr\_gsd \\%& 29.78  \\
  MS Arabic & ar\_padt  \\ %& 29.78  \\
%  Italian & it\_partut & 1781 & \\%& 29.78  \\
  English & en\_ewt  \\%& 29.78  \\
  Maltese & mt\_mudt &  \\%& 29.78  \\
  Vietnamese & vi\_vtb &  \\%& 29.78  \\
  %Chinese* & zh\_gsd &  \\%& 29.78  \\
\bottomrule
\end{tabular}
\caption[]{Universal Dependencies \cite{nivre2016universal} Datasets used for cross-lingual experiments}
\label{tab:sourcedata}
\end{table}

\subsection{Fine-tuning hyper-parameters}
\label{hyper}

We list here all the hyper-parameters used for fine-tuning in a supervised way on POS tagging and in an unsupervised way on raw \arz data (cf. Table \ref{tab:suphyperparameters} and \ref{tab:unsupervisedhyperparamters}).
For the supervised setting, we run a grid search on all the combination of hyper-parameters and select the best model on the validation set of the source language for both POS tagging and parsing.
\begin{table}[h!]
\footnotesize 
\centering
\begin{tabular}{lr}
\toprule
%  Parameter    & Value \\
%  \midrule
  %\arz &  & - &  \\
  batch size & \{32,16\} \\
  learning rate &  \{1e-5,5e-5,1e-4\}  \\
  optimizer & Adam \\%& 29.78  \\
  epochs (best of) &   30 \\%& 29.78  \\

  %learning rate warmup & linear & \\%& 29.78  \\
\bottomrule
\end{tabular}

\caption{Supervised fine-tuning hyper-parameters. }
\label{tab:suphyperparameters}
\end{table}

\iffalse
\begin{table}[h!]
\footnotesize
\centering
\scalebox{1}{
\begin{tabular}{lrr}
\toprule
   Parameter  & Value \\
  \midrule
  %\arz &  & - &  \\
  batch size & 64 \\
  learning rate &  5e-5  \\
  optimizer & Adam \\%& 29.78  \\
  warmup & linear \\%& 29.78  \\
  warmup steps & 10\% total \\%& 29.78  \\
  epochs (best of)  & 10 \\%& 29.78  \\
  %learning rate warmup & linear & \\%& 29.78  \\
\bottomrule
\end{tabular}
}
\caption{Unsupervised fine-tuning hyper-parameters}
% 
\label{tab:unsupervisedhyperparamters}
\end{table}
\fi

\begin{table}[h!]
\footnotesize
\centering
\begin{tabular}{lr}
\toprule
%   Parameter  & Value \\
%  \midrule
  %\arz &  & - &  \\
  batch size & 64 \\
  learning rate &  5e-5  \\
  optimizer & Adam \\%& 29.78  \\
  %learning rate warmup & linear & \\%& 29.78  \\
\bottomrule
\end{tabular}
~
\begin{tabular}{lr}
\toprule
%   Parameter  & Value \\
%  \midrule
  %\arz &  & - &  \\
  warmup & linear \\%& 29.78  \\
  warmup steps & 10\% total \\%& 29.78  \\
  epochs (best of)  & 10 \\%& 29.78  \\
  %learning rate warmup & linear & \\%& 29.78  \\
\bottomrule
\end{tabular}

\caption{Unsupervised fine-tuning hyper-parameters}
\label{tab:unsupervisedhyperparamters}
\end{table}

\subsection{Buckets : Detailed data and scores}

\begin{table}[h!]
\small\centering
\scalebox{0.9}{
\begin{tabular}{llrrr}
    \toprule
    Proportion Arabizi & $<$60 & 60-78 & 78-100 & =100  \\
    \% of word in sentence  & &  & &   \\
    \midrule
    train set number sents  & 322 & 286 & 283 & 276  \\
    %test set number sents  &  39 & 38 & 34 & 36  \\
%\% of total senteces   & 26 & 33 & 34 & 45  \\
    \bottomrule
\end{tabular}
}
\caption{Code-mixed Buckets}
\label{tab:buckets_code_mixed}
\end{table}

\iffalse
\begin{table}[h!]
\small\centering
\scalebox{0.8}{
\begin{tabular}{llrrrr}
    \toprule
    Proportion Arabizi  & $<$60  & 60-78 & 78-100 & =100  \\
    %Proportion Arabizi  & =100 & 100-  & 90-85 & 85-70 & \<40  \\
    \% of token in sentence  & &&&& \\
    \midrule
%    mBERT fr\_partut  & 26 & 33 & 34 & 45 &  60 \\
 %   \marbert fr\_partut  & 32 & 47 & 49 & 52.5& 62\\
    StanfordNLP  & 59.55 & 35.93 &  25.41 &  16.84   \\
 %   Majority Class + Punct  & 21.79 &  &  &  &   \\
%    mBERT arabizi  & 77 & 79 & 84 & 84 & 83 \\
%    \marbert arabizi  & 79 & 83 & 86 & 85 & 84 \\
%    \midrule
%StanfordNLP arabizi  & 82.22 & 82.35 & 81.97 & 86.27   & 87.68 \
    \bottomrule
\end{tabular}
}
\caption{Performance with regard to code-mixing proportion of baseline \stanford trained on the UD French Partut treebank}

\label{tab:details_code_mixed_scores}
\end{table}

\fi

%\bm{put the data in .zip appendix}

\end{document}